\pdfoutput=1
\documentclass[11pt]{article}

\usepackage[]{acl}
\usepackage{times}
\usepackage{latexsym}
\usepackage{enumitem}

\usepackage[T1]{fontenc}
\usepackage{CJKutf8}

\usepackage[utf8]{inputenc}
\usepackage[russian, english]{babel}

\usepackage{microtype}
\usepackage{graphicx}
\usepackage{tabu}
\usepackage{multirow}
\usepackage{arydshln}
\usepackage{times}
\usepackage{latexsym}
\usepackage{amssymb}
\usepackage{amsmath}
\usepackage{xspace} 
\usepackage{tablefootnote}
\usepackage{booktabs}
\usepackage{lingmacros}
\usepackage{colortbl}
\usepackage{subfig} %
\usepackage{todonotes}
\usepackage{soul}

\definecolor{CbFriendRed}{HTML}{D71B60}
\definecolor{CbFriendBlue}{HTML}{1E88E5}
\definecolor{CbFriendYellow}{HTML}{FFC107}
\definecolor{CbFriendGreen}{HTML}{004D3F}

\newcommand{\GenQA}{\textsc{GenQA}\xspace}
\newcommand{\astwo}{\textsc{AS2}\xspace}
\newcommand{\GenTyDiQA}{\textsc{Gen-TyDiQA}\xspace}
\newcommand{\MonoGenQA}{\textsc{Monolingual GenQA}\xspace}
\newcommand{\MultiGenQA}{\textsc{Multilingual GenQA}\xspace}
\newcommand{\CrossGenQA}{\textsc{Cross-Lingual GenQA}\xspace}
\newcommand{\MonoGenQAshort}{\textsc{MonoGenQA}\xspace}
\newcommand{\MultiGenQAshort}{\textsc{MultiGenQA}\xspace}
\newcommand{\CrossGenQAshort}{\textsc{CrossGenQA}\xspace}

\newcommand{\MTFIVE}{\textsc{mT5}\xspace}

    \title{Cross-Lingual Open-Domain Question Answering \\with Answer Sentence Generation}

\author{Benjamin Muller\textsuperscript{1}\thanks{\hspace{1em}Work conducted during internship at Amazon Alexa.}\hspace{.3em}, Luca Soldaini\textsuperscript{2}\thanks{\hspace{1em}Work conducted while employed at Amazon Alexa.}\hspace{.3em}, Rik Koncel-Kedziorski\textsuperscript{3}, Eric Lind\textsuperscript{3}, Alessandro Moschitti\textsuperscript{3}   \\
  \textsuperscript{1}Inria, Paris, France \\
  \textsuperscript{2}Allen Istitute for AI \\
  \textsuperscript{3}Amazon Alexa AI \\
  \texttt{benjamin.muller@inria.fr, lucas@allenai.org,}\\ \texttt{\{rikdz,ericlind,amosch\}@amazon.com}\\}

\begin{document}
\maketitle

\begin{abstract}

Open-Domain Generative Question Answering has achieved impressive performance in English by combining document-level retrieval with answer generation. These approaches, which we refer to as \GenQA, can generate complete sentences, effectively answering both factoid and non-factoid questions. In this paper, we extend \GenQA to the multilingual and cross-lingual settings. For this purpose,  we first introduce \GenTyDiQA, an extension of the TyDiQA dataset with well-formed and complete answers for Arabic, Bengali, English, Japanese, and Russian. Based on \GenTyDiQA, we design a cross-lingual generative model that produces full-sentence answers by exploiting passages written in multiple languages, including languages different from the question. Our cross-lingual generative system outperforms answer sentence selection baselines for all 5 languages and monolingual generative pipelines for three out of five languages studied.

\end{abstract}

\section{Introduction}

Improving coverage of the world's languages is essential for retrieval-based Question Answering (QA) systems to provide a better experience for non-English speaking users. 
One promising direction for improving coverage is multilingual, multi-source, open-domain QA.
Multilingual QA systems include diverse viewpoints by leveraging answers from multiple linguistic communities. 
Further, they can improve accuracy, as all facets necessary to answer a question are often unequally distributed across languages on the Internet \citep{Valentim}.

With the advance of large-scale language models, multilingual modeling has made impressive progress at performing complex NLP tasks without requiring explicitly translated data. 
Building on pre-trained language models \citep{devlin-etal-2019-bert,conneau-etal-2020-unsupervised,xue-etal-2021-mt5,liu-etal-2020-multilingual-denoising}, 
it is now possible to train models that accurately process textual data in multiple languages \citep{kondratyuk-straka-2019-75} and perform cross-lingual transfer  \citep{pires-etal-2019-multilingual} using annotated data in one language to process another language.

At the same time, answer generation-based approaches have been shown to be effective for many English QA tasks, including Machine Reading (MR) \cite{izacard2020leveraging,lewis2020retrieval}, question-based summarization \cite{iida2019exploiting,goodwin-etal-2020-towards,deng2020joint}, and, most relevant to this work, answer generation for retrieval-based QA \cite{DBLP:journals/corr/abs-2106-00955} --- that we refer to as \GenQA.

Compared to generative MR models, \GenQA approaches are trained to produce complete and expressive sentences that are easier to understand than extracted snippets \cite{Choi2021DecontextualizationMS}. %
Most importantly, they are trained to generate entire sentences, allowing them to answer both factoid or non-factoid questions, \text{e.g.}, asking for descriptions, explanation, or procedures. %

\begin{figure*}[h]
    \centering
    \includegraphics[height=3.3cm]{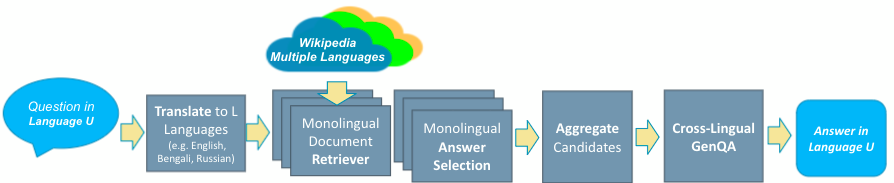}
    \caption{Illustration of our proposed Cross-Lingual, Retrieval-based \GenQA pipeline.}
    \label{fig:pipeline}

\end{figure*}
In this paper, we study and propose a simple technique for open-domain QA in a cross-lingual setting. Following \citet{DBLP:journals/corr/abs-2106-00955} (and as illustrated in Figure~\ref{fig:pipeline}), we work with a pipeline made of 3 main modules. First, a document retriever that retrieves relevant documents given a question; second, an answer sentence selection (AS2) model \citep{Garg_2020,vu2021multilingual} that ranks the sentences from the retrieved documents based on how likely they are to include the answer; and third, a generative model that generates a full sentence to answer the question given the sentence candidates. 

Our contribution focuses on the generative model. We introduce \CrossGenQAshort.  \CrossGenQAshort can generate full-sentence answers using sentence candidates written in multiple languages including languages different from the question and English.

Given the scarcity of annotated corpora for \GenQA, especially in languages different from English, we introduce the \GenTyDiQA dataset. \GenTyDiQA is an extension of TyDiQA, a dataset for typologically diverse languages in which questions are answered with passages and short spans extracted from Wikipedia  \citep{clark-etal-2020-tydi}. Our \GenTyDiQA includes human-generated, fluent, self-contained answers in Arabic, Bengali, English, Russian and Japanese, making it a valuable resource for evaluating multilingual generative QA systems. We found human-generated answers to be essential in evaluating \GenQA: compared to the standard approach of providing reference documents, they dramatically speed-up annotations and improve inter-annotator agreement.

Our evaluation shows that our \CrossGenQAshort system outperforms AS2 ranking models, and matches or exceeds similar monolingual pipelines. %

In summary, our contribution is three-fold:
\begin{enumerate}[topsep=0pt,itemsep=-1ex,partopsep=1ex,parsep=1ex,label=(\roman*)]
  \item We introduce \GenTyDiQA\footnote{We make \GenTyDiQA available at the following URL: \texttt{\small \url{s3://alexa-wqa-public/datasets/cross-genqa/}}}, an evaluation dataset that contains natural-sounding answers in Arabic, Bengali, English, Russian and Japanese, to foster the development of multilingual \GenQA systems.
  \item We confirm and extend the results of \citet{DBLP:journals/corr/abs-2106-00955} by showing that monolingual generative QA (\MonoGenQAshort) outperforms extractive QA systems in Arabic, Bengali, English and Russian.%
  \item We demonstrate that \CrossGenQAshort outperforms all our QA systems for Arabic, Russian, and Japanese, answering questions  using information from multiple languages.
\end{enumerate}

\section{Related Work}

\paragraph{Multilingual Datasets for QA}
Researchers have introduced several datasets for QA in multiple languages. Unlike our \GenTyDiQA, to the best of our knowledge, they are designed exclusively for extractive QA. \citet{Artetxe:etal:2019} extended the English machine reading SQuAD dataset \citep{rajpurkar-etal-2016-squad} by translating the test set to 11 languages.   Similarly, \citet{lewis-etal-2020-mlqa} collected new question and answer pairs for 7 languages following the SQuAD format.
Recently, \citet{mkqa} released MKQA, which includes question and answer pairs (predominantly Yes/No answers and entities) for 26 languages.
\citet{clark-etal-2020-tydi} released TyDiQA, a dataset for extractive QA in 11 typologically diverse languages.
 \citet{DBLP:journals/corr/abs-2010-12643} and \citet{shakeri-etal-2021-towards} have explored the use of techniques to synthetically generate data for extractive question answering using cross-lingual transfer. 

\paragraph{Generating Fluent Answers for QA}
The Generation of fluent and complete-sentence answers is still in its infancy, as most generative models for QA are used for extractive QA (e.g., \citep{Guu2020REALMRL,Lewis2020RetrievalAugmentedGF,asai-etal-2021-xor,DBLP:journals/corr/abs-2107-11976}. %
Approaches to ensure response fluency have been explored in the context of dialogue systems~\cite{baheti-etal-2020-fluent,Ni2021RecentAI}, but remain nevertheless understudied in the context of QA.
Providing natural sounding answers is a task of particular interest to provide a better experience for users of voice assistants.
One resource for this task is the MS-MARCO dataset \citep{DBLP:conf/nips/NguyenRSGTMD16}. 
It includes 182,669 question and answer pairs with human-written well-formed answers.
However, it only contains samples in English. 

Our \GenTyDiQA extends TyDiQA \citep{clark-etal-2020-tydi} adding natural human-generated answers for Arabic, Bengali, English, Japanese, and Russian. To the best of our knowledge, it is the first work that provides well-formed, natural-sounding answers for non-English languages.

\paragraph{Multilingual Extractive QA} Designing QA models for languages different from English is challenging due to the limited number of resources and the limited size of those datasets. 
For this reason, many studies leverage transfer learning across languages by designing systems that can make use of annotated data in one language to model another language. 
For instance, \citet{clark-etal-2020-tydi} showed that concatenating the training data from multiple languages improves the performance of a model on all the target languages for extractive QA.
In the Open-Retrieval QA setting, multilingual modeling can be used to answer questions in one language using information retrieved from other languages. 
\citet{10.1145/3077136.3080743} showed how cross-language tree kernels can be used to rank English answer candidates for Arabic questions.
\citet{DBLP:journals/corr/abs-2012-14094} designed a cross-lingual question similarity technique to map a question in one language to a question in English for which an answer has already been found.
\citet{asai-etal-2021-xor} showed that extracting relevant passages from English Wikipedia can deliver better answers than relying only on the Wikipedia corpora of the question language.
\citet{vu2021multilingual} showed how machine translated question-answer pairs can be used to train a multilingual QA model;
in their study, they leveraged English data to train an English and German AS2 model. 

Finally, \citet{cora} introduced CORA and reached state-of-the-art performance on open-retrieval span-prediction question answering across 26 languages. 
While related to our endeavor, it is significantly different in several key aspects.
First, unlike \CrossGenQAshort, CORA does not produce full, complete sentences; rather, it predicts spans of text that might contain a factoid answer.
Second, it mainly relies on sentence candidates that are written in English and in the question language; by contrast, in our work we choose to translate the questions into a variety of languages, allowing us to use monolingual retrieval pipelines to retrieve candidate sentences in diverse languages. We show that this form of cross-lingual \GenQA outperforms monolingual \GenQA in a majority of the languages studied.

\paragraph{Answer Sentence Selection (\astwo)}
The AS2 task originated in the TREC QA Track \cite{voorhees_2001}; more recently, it was revived by \citet{wang-etal-2007-jeopardy}. 
Neural AS2 models have also been explored \citep{DBLP:conf/iclr/Wang017,Garg_2020}. 
AS2 models receive as input a question and a (potentially large) set of candidate answers; they are trained to estimate, for each candidate, its likelihood to be a correct answer for the given question.

Several approaches for monolingual \astwo have been proposed in recent years.
\citet{severyn2015learning} used CNNs to learn and score question and answer representations, while others proposed alignment networks \citep{shen-etal-2017-inter,tran-etal-2018-context,10.1145/3219819.3220048}.
Compare-and-aggregate architectures have also been 
extensively studied \citep{DBLP:conf/iclr/Wang017,10.1145/3132847.3133089,DBLP:journals/corr/abs-1905-12897}. 
\citet{tayyar-madabushi-etal-2018-integrating} exploited fine-grained question classification to further improve answer selection.
\citet{Garg_2020} achieved state-of-the-art results by fine-tuning transformer-based models on a large QA dataset first, and then adapting to smaller AS2 dataset. 
\citet{matsubara-vu-2020-sigir} showed how, similar in spirit to \GenQA, multiple heterogeneous systems for \astwo can be be combined to improve a question answer pipeline.

\section{The \GenTyDiQA Dataset}
To more efficiently evaluate our multilingual generative pipeline (lower cost and higher speed), we built  \GenTyDiQA, an evaluation dataset for answer-generation-based QA in Arabic, Bengali, English, Japanese, and Russian. This extends the TyDiQA \citep{clark-etal-2020-tydi} dataset.

TyDiQA is a QA dataset that includes questions for 11 typologically diverse languages. Each entry is composed of a human-generated question and a single Wikipedia document providing relevant information. For a large subset of its questions, TyDiQA also contains a human-annotated passage extracted from the Wikipedia document, as well as a short span of text that answers the question.
We extend the TyDiQA validation set\footnote{The TyDiQA test set is not publicly available.} by collecting human-generated answers based on the provided questions and passages using Amazon Mechanical Turk\footnote{\url{https://requester.mturk.com}} (cf. Appendix~\ref{sec:mturk} for hiring criteria and rewards). %
Collecting human-generated answers is crucial for properly evaluating \GenQA models, as we will show in section~\ref{subsec:analysis}.
We use a two-stage data collection process:

\begin{table}[!t]
\centering\footnotesize
\renewcommand*{\arraystretch}{.99}
\resizebox{\linewidth}{!}{
\begin{tabular}{p{\linewidth}}

\toprule
\textbf{(\textsc{En}) Question}: What do pallid sturgeons eat? \\
\textbf{TyDiQA Span}: -- \\
\textbf{\GenTyDiQA Answer}: Pallid sturgeons eat various species of insects and fish depending on the seasons. \\
\midrule
\textbf{(\textsc{Ru}) Question}:
\begin{otherlanguage*}{russian}
 Когда закончилась Английская революция?
\end{otherlanguage*}\small{\textit{When did the English Revolution end?}}\\
\textbf{TyDiQA Span}: 1645 \\
\textbf{\GenTyDiQA Answer}:
\begin{otherlanguage*}{russian}Английская революция, известная также как Английская гражданская вой закончилась в 1645, когда Кромвель создал «Армию нового образца», одержавшую решающую победу в сражении при Нэйcби
\end{otherlanguage*}\small{\textit{The English Revolution, also known as the English Civil War; ended in 1645, when Cromwell created the "Army of the new model", which won a decisive victory at the Battle of Naysby.}}\\
\midrule
{\begin{CJK}{UTF8}{min}\textbf{(\textsc{Ja)} Question}: ストーンズリバーの戦いによる戦死者は何人 \small{\textit{How many were the deaths from the Battle of Stones River?}}\end{CJK}}\\
{\begin{CJK}{UTF8}{min}\textbf{TyDiQA Span}: 23,515名 \textit{23,515 people}\end{CJK}}\\
{\begin{CJK}{UTF8}{min}\textbf{\GenTyDiQA Answer}: ストーンズリバーの戦いで23,515人が川で殺されました。 \small{\textit{23,515 people were killed in the river in the Battle of Stones River.}}\end{CJK}}\\
\bottomrule
\end{tabular}
}
\caption{\GenTyDiQA question and answer samples.}%
\label{table:gentydiqa}
\vspace{-1em}
\end{table}

\paragraph{(1) Answer Generation}
We show each turker a question and its corresponding passage, and ask them to write an answer that meets the following three properties:
(\textit{i}) The answer must be \textbf{factually correct and aligned} with the information provided in the passage. 
If a passage is not sufficient to answer a question, turkers will respond ``no answer''.
(\textit{ii}) The answer must be a \textbf{complete and grammatically correct} sentence, or at most a few sentences.
(\textit{iii}) The answer should be \textbf{self-contained}; that is, it should be understandable without reading the question or the passage. 
Based on this condition, ``yes'' or ``no'' are not acceptable answers.

\paragraph{(2) Answer Validation}
We show each question alongside its corresponding passage and the human-generated answer from Step (1) to five turkers. We ask them to label if the collected answer meets the three properties listed above: correctness, completeness, and self-containedness. We aggregate labels and keep only answers that received at least 3/5 positive judgements for each property. 
Table~\ref{table:gentydiqa} contains some examples of the data collected.

\paragraph{Data Statistics}
We report the number of \GenTyDiQA collected human-generated natural answers in table~\ref{tab:gentydiqa}, and our coverage of the TyDiQA dataset.
We do not reach 100\% coverage due to our highly selective validation stage: we only accept answers that receive 3/5 votes for each property, a process that guarantees a high-quality dataset.

\section{Multilingual GenQA Systems}
Our goal is to build a QA system that, given a question in a target language, retrieves the top-$k$ most relevant passages from text sources in multiple languages, and generates an answer in the target language from these passages (even if the passages are in a different language from the question).

\begin{table}[t!]
\centering\footnotesize
\renewcommand*{\arraystretch}{.9}
\resizebox{\linewidth}{!}{
\begin{tabu}{ l   @{\hspace{0.35cm}} c  c c}%
	\toprule
	Lang. (iso)
	& 
	\#Answers  &  Avg. Length (utf-8)  & \%TyDiQA\\ 

	\midrule
	Arabic ~~(\textsc{Ar}) &  859 &  152.5 & 75.7      \\ %
	Bengali ~(\textsc{Bn}) &    89    &   177.2  &  63.6 \\
	English ~(\textsc{En}) &    593    &  64.0  &  79.5   \\ %
	Japanese (\textsc{Ja}) &  550  &112.0 &  62.1\\ %
 
	Russian ~(\textsc{Ru}) & 595 & 277.9 &  52.6 \\ %

	\bottomrule
\end{tabu}
}
\caption{\small{Statistics on \GenTyDiQA Answers}}%
\label{tab:gentydiqa}
\vspace{-1em}
\end{table}

\subsection{Task Definition and System Architecture}

We first describe the AS2 and \GenQA tasks in a language-independent monolingual setting, and then generalize to the cross-lingual setting.

In the monolingual setting for a language $L_i$, an AS2 system takes as input a question $q$ and a possibly large set of candidate answers $C_{L_i}$ (e.g. all sentences from Wikipedia in the language $L_i$), ranks each candidate answer given $q$, and returns the top-ranking candidate $c_m \in C_{L_i}$. A \GenQA system uses the top $k$ AS2-ranked answers in $C_{L_i}$ to synthesize a machine-generated answer $g$ in language $L_i$. 

The cross-lingual \GenQA task extends this setup as follows: Consider a set of languages $\{L_1,\dots,L_r\}$. Given a question $q$ in language $L_i$, let $M = \cup_{j=1}^{r} C_{L_j}$ be the set of relevant candidate sentence answers for $q$ in any language. A cross-lingual \GenQA system uses the top $k$ ranked answers in $M$ --- regardless of language --- to generate an answer $g$ in $L_i$.

Our architecture, illustrated in Figure~\ref{fig:pipeline}, consists of the following components: (\textit{i}) question translation\footnote{We used Amazon's AWS Translate service, \texttt{\url{https://aws.amazon.com/translate/service}}. We validate the quality of AWS Translate on the languages we study in the Appendix section~\ref{sec:translate_aws}.} from $L_i$ to produce queries $q_{L_j}$ in each language $L_j$, (\textit{ii}) a document retriever for each $L_j$ to get $C_{L_j}$, (\textit{iii}) a monolingual AS2 model for each language, which sorts the candidates in $C_{L_j}$ in terms of probability to be correct given $q_{L_j}$, where $C_{L_j}$ is created by splitting the retrieved documents into sentences, (\textit{iv}) an aggregator component, which builds a multilingual candidate set $M$ using the top $k$ candidates for each language, and (\textit{v}) a cross-lingual answer generation model, which generates $g$ from $M$.

We now present in more details each component of our system.

\subsection{Multilingual Passage Retrieval}
\label{sec:retriever}
To obtain candidates for our multilingual pipeline, we used Wikipedia snapshots collected in May 2021. 
We processed each snapshot using WikiExtractor   \citep{Wikiextractor2015}, and create monolingual indices using PyTerrier \citep{pyterrier2020ictir}. 
During retrieval, we first translate queries in each language using AWS Translate. We validate the good quality of this system for all our languages in table~\ref{tab:aws_translate_score} in the Appendix. We then use BM25 \cite{robertson1995okapi} to score documents.
We choose BM25 because, as shown by \citet{thakur2021beir}, it is competitive with DPR-based models \citep{karpukhin-etal-2020-dense} and it outperforms DPR across a great diversity of domains. 

\paragraph{Evaluation}
We evaluate the different retrievers independently: for each question, we compare the exact match of the title of the retrieved document with the gold document's title provided by TyDiQA. We compute the Hit@N at the document level, i.e., the percentage of questions having a correct document in the top-N predicted documents. In our experiments, we retrieve the top-100 documents from Wikipedia to feed them to the AS2 model.

\subsection{AS2 models for different languages}
\label{sec:as2}
We build AS2 models by fine-tuning the multilingual masked-language model XLM-R  \cite{conneau-etal-2020-unsupervised} into multiple languages, using question/sentence pairs, which we created with the TyDiQA dataset.
We followed the procedure by \citet{Garg_2020} performed on the NQ dataset \citep{kwiatkowski-etal-2019-natural} to build the ASNQ dataset for English. 
For each $\langle\text{question}, \text{Wikipedia document}, \text{span}\rangle$ triplet from the TyDiQA dataset, we use the span to identify positive and negative sentence candidates in the Wikipedia document. We first segment each document at the sentence level using the \verb|spacy| library\footnote{\url{https://spacy.io/}}.
We define positive examples to be the sentences that contain the span provided by the TyDiQA dataset, and negative examples to be all other sentences from the same Wikipedia document.  We report statistics on  AS2-TyDiQA in the Appendix in table~\ref{tab:as2tydiqa}.
For more details, we refer the reader to \citet{Garg_2020}.

\paragraph{Model}
We fine-tune XLM-R extended with a binary classification layer on the AS2-TyDiQA dataset described above. At test time, we rank the candidates using the model output probability. %
Preliminary experiments confirmed the results of \citet{clark-etal-2020-tydi} regarding machine reading models on TyDiQA : the best performance is obtained when concatenating the datasets from all languages.

\subsection{Multilingual Answer Generation Models}
\label{multilan}
We extended the work of \citet{DBLP:journals/corr/abs-2106-00955} on monolingual \GenQA modeling.  For each question, this model takes the top-5 candidates ranked by AS2 as input.
For \CrossGenQA, we build a set of multiligual candidates $M$ with two methods: (i)~\textsc{Top~2~/~lang.}, which selects the top 2 candidates for each language and concatenates them (in total $2\times5=10$); and (ii)~\textsc{Top~10}, which selects the 10 candidates associated with the highest scores regardless of their language. %

\paragraph{Model}
We used the pre-trained multilingual T5 language model (\MTFIVE) by \citet{xue-etal-2021-mt5}. This is an encoder-decoder transformer-based model \citep{NIPS2017_3f5ee243} pre-trained with a span-masking objective on a large amount of web-based data from 101 languages (we use the base version).
We fine-tuned \MTFIVE following \citep{DBLP:journals/corr/abs-2106-00955}: for each sample, we give the model the question concatenated with the candidates $M$ as input and a natural answer as the generated output. 
 \GenQA models are trained on MS-MARCO \citep{DBLP:conf/nips/NguyenRSGTMD16}\footnote{Using the train split of the NLGEN(v2.1) version.}, which includes 182,669 examples of $\langle\text{question}, \text{10 candidate passages}, \text{natural answer}\rangle$ instances in English. 
 When the language of the question (and answer) is not English or when we use candidates in multiple languages, we translate the training samples with Amazon's AWS Translate service and fine-tune the model on the translated data. 
 For instance, to design a \GenQA model answering questions in Arabic using input passages in Arabic, English, and Bengali, we fine-tune the model with questions and gold standard answers translated from English to Arabic, and input candidates in English, Arabic, and Bengali, where the latter two are translated from the MS-MARCO English passages.%

\paragraph{Evaluation}
\label{sec:evaluation_genqa}

As pointed out by \citet{chen-etal-2019-evaluating},  automatically evaluating generation-based QA systems is challenging. We experimented with BLEU~\citep{papineni-etal-2002-bleu} and ROUGE-L \citep{lin-2004-rouge}, two standard metrics traditionally used for evaluating generation-based systems, but found that they do not correlate with human judgment. For completeness, we report them in the Appendix~\ref{sec:evaluation_bleu_vs_accuracy} along with a detailed comparison with human judgment. Thus, we rely on human evaluation through Amazon Mechanical Turk\footnote{We describe in \ref{sec:mturk} how we choose and reward turkers.}: we ask three turkers to vote on whether the generated answer is correct, and report the $\frac{\sum PositiveVotes}{\sum Total Votes}$ as system Accuracy.

\section{Experiments}

Multilinguality and the different components of our system pipeline raise interesting research questions.  Our experimental setup is defined by the combinations of our target set of languages with respect to questions, candidates, and answers.
We experiment with \GenQA in the monolingual (one model per language) and multilingual (one model for several languages) settings, where the question and candidates in the same language are used to generate an answer. Then we experiment with a cross-lingual \GenQA model that is fed candidates in multiple languages. Despite being an apparent more complex task, we find that in many cases, the cross-lingual model outperform all other settings.

\subsection{Setup}
We approach multilingual generation-based question answering in three ways:
\paragraph{\MonoGenQA (\MonoGenQAshort)}
The candidate language is the same as the question. For each language (Arabic, Bengali, English, Japanese and Russian), we monolingually fine-tune \MTFIVE, and report the performance of each \GenQA model on the \GenTyDiQA dataset (Tab.~\ref{tab:cross_lingual_candidates}).

Our contribution is to show that this approach, first introduced by \citet{DBLP:journals/corr/abs-2106-00955} for English, delivers similar performance for other languages.

\paragraph{\MultiGenQA (\MultiGenQAshort)}
We train one \MTFIVE for all five languages by concatenating their training and validation sets. This single model can answer questions in multiple languages, but it requires that answer candidates be in the same language as the question. We report the performance of this \MultiGenQAshort model in table~\ref{tab:cross_lingual_candidates}.

For this set of experiments, we show that a single multilingual \GenQA model can compete with a collection of monolingual models.

\paragraph{\CrossGenQA (\CrossGenQAshort)}
We use candidates in multiple languages (Arabic, Bengali, Russian, English, Arabic) to answer a question in a target language. We retrieve and rerank sentence candidates in each language, aggregate candidates across all the languages, and finally generate answers (in the same language as the question). We report the performance on the \GenTyDiQA dataset (table~\ref{tab:cross_lingual_candidates}).

These experiments aim to determine whether our generative QA model can make use of information retrieved from multiple languages and outperform the baseline methods.

\paragraph{Manual Evaluation}

We stress the fact that all the results derived in the following experiments were manually evaluated with Amazon Mechanical Turk. In total, we run 34 tasks (system evaluations), requiring around 60k Hits, for a total manual evaluation of 20k QA pairs (times 3 turkers). %

\begin{table}[t!]
\centering\footnotesize
\renewcommand*{\arraystretch}{.9}
\resizebox{5.5cm}{!}{
\begin{tabu}{ l l  c}%
	\toprule
	\textbf{Model} & \textbf{\textsc{Candidates}}  & \textbf{Accuracy}
	\\
	\midrule
\MonoGenQAshort	& \textsc{En} & \bf 77.9   \\ %
\CrossGenQAshort	&	\textsc{De}  & \underline{70.5} \\ %
\CrossGenQAshort	&\textsc{De Es Fr It} & 68.8 \\ %
\CrossGenQAshort	&	\textsc{Ar Ja Ko}  & 31.4\\ %
Clozed-Book	&\textsc{None}  & 21.0 \\ %
	
\bottomrule
\end{tabu}
}
\caption{Impact of the candidate language set on \CrossGenQA in English on MS-MARCO. %
The language set is controlled with machine translation.
}%
\label{tab:translated_candidates}
\vspace{-1em}
\end{table}

\subsection{Feasibility Study}

To explore whether a model fed with candidates written in languages different from the question can still capture relevant information to answer the question, we conduct a feasibility study using the MS-MARCO dataset with English as our target language and machine translated candidates. 

For each question, we translate the top 5 candidate passages to different languages and provide these translated candidates as input to the model. We experiment with three translation settings: all candidates translated to German (\textsc{De}); each candidate translated to a random choice of German, Spanish, French or Italian (\textsc{De-Es-Fr-It}); translated to  Arabic, Japanese or Korean (\textsc{AR-JA-KO}). We compare all these \CrossGenQA models with a Clozed-Book QA Model \citep{roberts-etal-2020-much} for which no candidates are fed into the model.

\paragraph{Results} We report the performance in table~\ref{tab:translated_candidates}.  All \CrossGenQA models outperform significantly the Clozed-book approach. This means that even when the candidates are in languages different from the question, the model is able to extract relevant information to answer the question. We observe this even when the candidates are in languages distant from the question language (e.g., Arabic, Japanese, Korean).

\begin{table}[t!]
\centering\small
\resizebox{\linewidth}{!}{
\begin{tabu}{ l   @{\hspace{0.35cm}} l    c c}%
	\toprule
	
	\bf Language  &  \bf  BLEU   & \bf  ROUGE & \bf  Accuracy
	\\%

	\midrule
    \textit{\MonoGenQA} &\\
    \textsc{Ar} &    24.8 / 17.2  &  47.6 / 38.8 & 77.1 / 68.4  \\ %
    \textsc{Bn}&    27.4 / 21.7 &  48.6 / 43.0 & 82.0 / 67.4 \\

     \textsc{En} &  31.5 / 23.0  &    54.4 / 46.4 &  68.5 / 43.6 \\
     \textsc{Ja}  &   24.5 / 19.4  & 50.2 / 45.0  & 72.3 / 64.3 \\ 
    \textsc{Ru}  &   10.2 / 6.4 &  30.2 / 23.4 & 82.6 / 61.3  \\ %

	\midrule
	\textit{\MultiGenQA} &      \\
    \textsc{Ar}  &   24.3 / 17.4 & 47.9 / 39.0  &  74.9 /  72.7 \\ %
	\textsc{Bn}  &     27.3 /  23.7  &  47.8 / 44.9 &  84.3 /  76.5 \\ %

	\textsc{En}  &    30.8 / 21.8 &  54.5 / 46.2   &   65.3 / 37.4
	  \\
	\textsc{Ja}  &   23.9 / 19.1  &  50.0 / 45.5  &  76.8 / 65.5  \\ %
	\textsc{Ru}  &   10.6 /  6.4  &  31.0 / 23.2   & 76.6 /  66.7 \\

	\bottomrule
\end{tabu}
}
\caption{Performance of our \GenQA models fine-tuned on MSMARCO and evaluated on \textsc{GenTyDiQA} using Gold-Passage from TyDiQA/Ranked Candidates from Wikipedia.
}%
\label{tab:multilingual_gen_QA}
\end{table}

\subsection{\GenTyDiQA Experiments}

This section reports experiments of the full \GenQA pipeline tested on the \GenTyDiQA dataset with candidates retrieved from Wikipedia. For each question, we retrieve documents with a BM25-based retriever, rank relevant candidates using the AS2 model, and feed them to the \GenQA models. We note that we cannot compare the model performance across languages: as pointed out in \citep{clark-etal-2020-tydi} regarding TyDiQA. %

\paragraph{\MonoGenQAshort Performance}
We measure the impact of the retrieval and AS2 errors by computing the ideal \GenQA performance, when fed with gold candidates (TyDiQA gold passage). We report the results in table~\ref{tab:multilingual_gen_QA}. 
\label{sec:monolingual_candidates}
We evaluate the performance of the \GenQA models, also comparing it to AS2 models on the \GenTyDiQA dataset of each language. We report the results in table~\ref{tab:cross_lingual_candidates} (cf. \MonoGenQAshort). The first row shows the document retrieval performance in terms of Hit@100 for the different languages considered in our work. We note comparable results among all languages, where Arabic reaches the highest accuracy, 70.7, and Japanese the lowest, 57.0. %
The latter may be due to the complexity of indexing ideogram-based languages. However, a more direct explanation is the fact that retrieval accuracy strongly depends on the complexity of queries (questions), which varies across languages for \GenTyDiQA. %
Similarly to \citet{clark-etal-2020-tydi}, we find that queries in English and Japanese are more complex to answer compared to other languages.

Regarding answering generation results,  rows 2 and 3 for English confirm \citet{DBLP:journals/corr/abs-2106-00955}'s findings: \GenQA outperforms significantly AS2 by 4.6\% (43.6 vs. 39.0). We also note a substantial improvement for Bengali (+9.4\%, 67.4 to 58.0). In contrast, Arabic and Russian show similar accuracy between \GenQA and AS2 models.
Finally, AS2 seems rather more accurate than  \GenQA for Japanese (70.4 vs 64.3). 
Results reported by \citet{xue-etal-2021-mt5} show \MTFIVE to be relatively worse for Japanese than all other languages we consider in many downstream tasks, so the regression seen here might be rooted in similar issues.

\newcommand{\allCandidates}{\textsc{All}\xspace}

\begin{table}[t!]
\centering\footnotesize
\renewcommand*{\arraystretch}{.9}
\resizebox{\linewidth}{!}{
\begin{tabu}{l c c c c c}%
	\toprule
	
    \textbf{Model}	& \textbf{\textsc{Ar}} & \textbf{\textsc{Bn}} & \textbf{\textsc{En}} & \textbf{\textsc{Ja}} & \textbf{\textsc{Ru}}
	\\
	\midrule
    \textsc{Retriever} (Hit@100 doc.)           & 70.7  & 66.3 & 66.9 & 57.0 & 67.8 \\
    \midrule
                AS2                             & 68.0 & 58.0 & \underline{39.0} & \underline{70.4} & 60.8 \\
    \MonoGenQAshort                             & 68.4  & \underline{67.4} & \bf 43.6 & 64.3 & 61.3 \\
    \MultiGenQAshort                            & \underline{72.7}  & \bf 76.5 & 37.4 & 65.5 & \underline{66.7}
    \\
    \CrossGenQAshort\textsc{top 10}              & 72.0 & 25.3  & 31.0 & \underline{70.3} & \bf{74.3}
    \\
    \CrossGenQAshort\textsc{top. 2~/~lang.}     & \bf 73.2 & 18.5  & 29.3 & \bf {71.6} & \bf 74.7
    \\

	\bottomrule
\end{tabu}
}
\caption{Hit@100 doc.~of the retriever and Accuracy of \GenQA models on \GenTyDiQA. All \CrossGenQAshort experiments use candidates aggregated from all the languages (\textsc{Ar, Bn, En, Ja, Ru}).
}%
\label{tab:cross_lingual_candidates}
\vspace{-1em}
\end{table}

\paragraph{\MultiGenQAshort Performance}
\label{sec:multilingual_pipeline}
We compare the performance of the \MonoGenQA models (one model per language) to the performance of the \MultiGenQA model fine-tuned after concatenating the training datasets from all the languages. We report the performance in table~\ref{tab:cross_lingual_candidates} (cf. \MultiGenQAshort): multilingual fine-tuning improves the performance over monolingual fine-tuning for all languages except English. This shows that models benefit from training on samples from different languages. For Bengali, we observe an improvement of around 9\% in accuracy. 
This result has a strong practical consequence: at test time, we do not need one \GenQA model per language, we can rely on a single multilingual model trained on the concatenation of datasets from multiple languages (except for English, where we find that the monolingual model is more accurate). This result generalizes what has been shown for extractive QA \citep{clark-etal-2020-tydi} to the \GenQA task.

\paragraph{\CrossGenQAshort Performance}
\label{sec:cross_lingual_candidates}
Our last and most important contribution is in table~\ref{tab:cross_lingual_candidates}, which reports the performance of a \GenQA model trained and evaluated with candidates in multiple languages. This model can answer a user question in one language (e.g., Japanese) by using information retrieved from many languages, e.g.,  Arabic, Bengali, English, Japanese, and Russian). 
For Arabic, Japanese, and Russian, we observe that \CrossGenQA outperforms other approaches by a large margin, e.g., for Russian, 13.8\% (74.6-60.8) better than AS2, and an 8\% percent improvement over \MultiGenQAshort. 

For Bengali, the model fails at generate good quality answers (\CrossGenQAshort models reach at best 25.3\% in accuracy compared to the 76.9\% reached by the \MultiGenQAshort model). We hypothesize that this is the consequence of a poor translation quality of the question from Bengali to other languages such as English, Arabic, or Japanese, which leads to poor candidate retrieval and selection, ultimately resulting in inaccurate generation.

\begin{table}[t!]
\centering\footnotesize
\renewcommand*{\arraystretch}{.9}
\resizebox{\linewidth}{!}{
\begin{tabu}{ l     l c}%
	\toprule
	
    \textbf{Model}	& \textbf{Candidates}  & \textbf{Accuracy}
	\\
	\midrule

	\MonoGenQAshort & \textsc{En} &   57.8 \\ %
	\CrossGenQAshort & \textsc{Ja} &   \underline{60.3} \\
    \CrossGenQAshort &  \textsc{Ar-Bn-En-Ja-Ru} \textsc{top 10}  &   56.9  \\
	   \CrossGenQAshort &  \textsc{Ar-Bn-En-Ja-Ru} \textsc{top 2 / lang} &   \bf 63.8 \\

	\bottomrule
\end{tabu}
}
\caption{\GenQA scores in English on Japanese-culture-specific questions extracted from TyDiQA. \textsc{Candidates} defines the language set of the input candidates.
}%
\label{tab:culture_specific_qa}
\end{table}

Finally, we compare the two candidate aggregation strategies used for \CrossGenQA: \textsc{Top~2~/~lang.} and \textsc{Top~10} (see section~\ref{multilan}).
We observe that the aggregation strategy impacts moderately the downstream performance. For English, Arabic, Japanese and Russian the gap between the two methods is at most 2 points in accuracy. %
We leave the refinement of  candidate selection in the multilingual setting for future work. %

\subsection{Analysis}
\label{subsec:analysis}

\paragraph{Examples}%
Table~\ref{table:prediction_example_ru} shows the output of AS2,  \MultiGenQA, and \CrossGenQA models to questions in Russian and Bengali. 
For Bengali, the \GenQA models provide a correct and fluent answer while the AS2 model does not. For Russian, only the \CrossGenQA model is able to answer correctly the question. This because AS2 does not rank the right information in the top k, while \CrossGenQA can find the right information in another language in the multi-language candidate set.

\begin{table}[t]
\renewcommand*{\arraystretch}{.9}
\centering\footnotesize
\resizebox{\linewidth}{!}{
\begin{tabular}{p{\linewidth}}

\toprule

\textbf{Question}:
\begin{minipage}{.27\textwidth}
      \includegraphics[width=\linewidth]{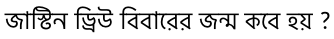}
\end{minipage}

\small{\textit{When was Justin Drew Bieber born?}}\\
\textcolor{purple}{\textbf{AS2 Prediction:}}\\
\begin{minipage}{.48\textwidth}
      \includegraphics[width=\linewidth]{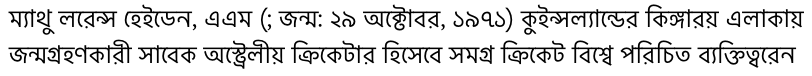}
\end{minipage}
\\
\small{\textit{
 Matthew Lawrence Hayden, AM (born October 29, 1971) is a former Australian cricketer born in Kingroy, Queensland.}}
\\
\textcolor{CbFriendBlue}{\textbf{\MultiGenQAshort Prediction:}}\\
\begin{minipage}{.45\textwidth}
      \includegraphics[width=\linewidth]{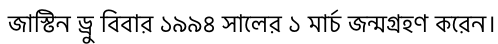}
\end{minipage}
\small{\textit{Justin Drew Bieber was born on March 1, 1994.}}\\
\textcolor{CbFriendBlue}{\textbf{\CrossGenQAshort Prediction}}\\
\begin{minipage}{.45\textwidth}
      \includegraphics[width=\linewidth]{imgs/bn_genqa_multi_cross.png}
\end{minipage}

\small{\textit{Justin Drew Bieber was born on March 1, 1994.}}
\\
\midrule

{\begin{CJK}{UTF8}{min}\textbf{Question}: トゥールのグレゴリウスはいつ生まれた？
\small{\textit{When was Gregory of Tours born?
}}\end{CJK}}\\

{\begin{CJK}{UTF8}{min}\textcolor{purple}{\textbf{AS2 Prediction:}} グレゴリウス14世（Gregorius XIV,1535年2月11日 - 1591年10月16日）はローマ教皇（在位：1590年 - 1591年）。
\small{\textit{Pope Gregory XIV (February 11, 1535 – October 16, 1591) is the Pope of Rome (reigned: 1590 – 1591).
}}\end{CJK}}\\
{\begin{CJK}{UTF8}{min}\textcolor{purple}{\textbf{\MultiGenQAshort Prediction:}}トゥールのグレゴリウスは、1535年2月11日に生まれた。
\small{\textit{Gregory of Tours was born on February 11, 1535.}}\end{CJK}}\\
{\begin{CJK}{UTF8}{min}\textcolor{CbFriendBlue}{\textbf{\CrossGenQAshort Prediction}}トゥールのグレゴリウスは538年頃11月30日に生まれた。 \small{\textit{Gregory of Tours was born on November 30, 538.}}\end{CJK}}\\
\bottomrule
\end{tabular}
}
\caption{Example of predicted answers to questions in Bengali and Japanese. \textcolor{CbFriendBlue}{\textbf{Blue}} indicates correct predictions while \textcolor{CbFriendRed}{\textbf{Red}} incorrect ones.
Translations are intended for the reader and are not part of the predictions.}

\label{table:prediction_example_ru}
\end{table}

\paragraph{Error Propagation}
We observe (table~\ref{tab:multilingual_gen_QA}) that the \GenQA models are highly impacted by the retriever and AS2 quality. For example, English \GenQA performance drops of 27.9 (65.3-37.4) points in Accuracy. This suggests that large improvement could be achieved by improving the document retriever and/or AS2 modules.

\begin{table}[b]
\centering\footnotesize
\renewcommand*{\arraystretch}{.9}
\begin{tabular}{lccc}
\toprule
\textbf{Eval mode} &
  \begin{tabular}[c]{@{}c@{}}
  \textbf{Strong} \\ \textbf{agreement}\end{tabular} &
  \begin{tabular}[c]{@{}c@{}}\textbf{Perfect} \\ \textbf{agreement}\end{tabular} &
  \begin{tabular}[c]{@{}c@{}}\textbf{Fleiss'}\\ \textbf{kappa}\end{tabular} \\
\midrule
No Reference &
  55.00 \% &
  16.43 \% &
  0.1387 \\
With Reference &
  85.36 \% &
  55.25 \% &
  0.5071 \\
 \bottomrule
\end{tabular}
\caption{Comparison between providing a reference answer and not for evaluating \MonoGenQAshort predictions~(\textsc{En}).
Providing a reference increases agreement.}
\label{tab:ref_noref}
\end{table}

\paragraph{Culture-Specific Questions in English}
One striking result across our experiments is the lower performance of \CrossGenQA model than \GenQA model on English. We hypothesize that English questions from the \GenTyDiQA dataset are more easily answered using information retrieved from English compared to other languages because those questions are centered on cultures specific to English-speaking countries. 

To verify our hypothesis, we re-run the same set of experiments, using culture-specific Japanese questions rather than English queries. To do so, we (i) took the Japanese questions set from \GenTyDiQA, (ii) manually translated it in English, (iii) manually select 116 questions that are centered on Japanese culture, and (iv) run the same \GenQA pipeline on those questions. The results reported in table~\ref{tab:culture_specific_qa} show that \CrossGenQAshort outperforms \MonoGenQAshort, suggesting that the former improves also the English setting if the question set is culturally not centered on English, i.e., it requires answers that cannot be found in English.

\paragraph{Use of Reference Answer in Model Evaluation} 
We found the use of human-generated reference answers to be crucial to ensure a consisted annotation of each model. 
A comparison between annotation with and without reference answer is provided in table~\ref{tab:ref_noref}.
When using a reference, we found annotators to be dramatically more consistent, achieving a Fleiss' Kappa \citep{Fleiss1971MeasuringNS} of $0.5017$;
when providing no reference answer, the inter-annotation agreement dropped to $0.1387$. 
This trend is reflected in the number of questions with strong (4+ annotators agree) and perfect agreement.

\section{Limits}

Our system requires translating the questions. We also use the standard BM25 approach. Even though it was shown to be more robust compared to dense retriever \citep{thakur2021beir,Rosa2022NoPL}, using a cross-lingual retriever \citep{Li2021LearningCI} could improve performance and save the cost of translating the question. This has been explored by \citet{cora} but their retriever mainly retrieves passages in English and the question language %
which may lead to English-centric answers. Another limit is the fact that our system is not designed to handle questions that are not answerable. In the future, we may want to integrate a no-answer setting to avoid unwanted answer.

\section{Conclusion}

We study retrieval-based Question Answering systems using answer generation in a multilingual context. We proposed (i) 
\GenTyDiQA, a new multilingual QA dataset that includes natural and complete answers for Arabic, Bengali, English, Japanese, and Russian; based on this dataset (ii) the first multilingual and cross-lingual \GenQA retrieval-based systems. The latter can accurately answer questions in one language using information from multiple languages, outperforming answer sentence selection baseline for all languages and monolingual pipeline for Arabic, Russian, and Japanese.

\bibliography{anthology,custom}
\bibliographystyle{acl_natbib}

\clearpage

\appendix
\section{Discussion}
\label{sec:appendix_discussion}
\subsection{Machine Translation of the Questions and BM25 Retriever Engines}
Our work introduces \CrossGenQA, a system that can answer questions --- with complete sentence answers --- in multiple languages using candidates in multiple languages, possibly distinct from the question. They were many possible design choices to achieve such a goal. We chose to rely on automatically translating the questions before retrieving relevant documents in several languages using multiple (monolingual) BM25 retrievers. We could have chosen to use the recently released multilingual Dense passage Retrieval (mDPR) \citep{DBLP:journals/corr/abs-2107-11976}. We decided not to for the two following reasons. First, as shown by \citet{thakur2021beir}, BM25 is a very reasonable design choice for a retriever engine, that outperforms other approaches in many settings (including dense retrievers). Second, as seen in \citep{DBLP:journals/corr/abs-2107-11976}, multilingual dense retrievers usually retrieve passages in the same language as the question or English. This means that mDPR is highly biased toward the English language. In our work, by combining translation and monolingual retrievers, we can control the language set that we use for answer generation. We leave for future work the refinement of mDPR to enable for more diversity in the retrieved passage languages and to integrate it in our pipeline.

\subsection{Machine Translation Errors}

\begin{table}[b!]
\footnotesize\small 
\renewcommand*{\arraystretch}{1.15}
\centering
\resizebox{\linewidth}{!}{
\begin{tabular}{lccccc}
\toprule
  & ar & bn & en & ja & ru  \\
  \hline
  ar &  & \textbf{25.9}/16.1 & \textbf{40.8}/25.5 & \textbf{26.1}/16.0 & \textbf{27.3}/17.8 \\
  bn & \textbf{22.8}/10.7 & & \textbf{32.8}/22.9 & \textbf{23.5}/16.5 &  \textbf{21.8}/14.7  \\
  en & \textbf{39.5}/17.9 & \textbf{32.7}/23.0 & & \textbf{34.2}/22.8 & \textbf{36.6}/27.1\\
  ja & \textbf{21.0}/10.3 & \textbf{22.6}/16.0 & \textbf{28.0}/19.4 & & \textbf{21.4}/15.3  \\
  ru & \textbf{25.9}/13.5 & \textbf{24.9}/18.1 & \textbf{37.3}/27.5 & \textbf{26.4}/20.3 &  \\

\bottomrule
\end{tabular}}
\caption{Performance measured with spBLEU of AWS translate compared to a Many-to-Many (M2M) Multilingual Transformer Model (reported in \citep{Goyal2022TheFE}) on the FLORES devtest dataset \citep{Goyal2022TheFE}. Cell($i$,$j$) reports the score of AWS/M2M from language $i$ to language $j$. AWS translate outperforms the M2M model for all language pairs.  
}
\label{tab:aws_translate_score}
\end{table}

At test time, our system applies Machine Translation to the question to formulate queries in different languages and retrieve candidates for these languages using the BM25 retrieval engine. To our knowledge this is the best approach to generate queries in different languages, as MT systems are very powerful tools, trained on millions of data points and, thanks to Transformer model, they take the entire question context into account (other cross-query formulations can be applied but they will be probably less accurate and multilingual DPR is an excellent research line but not as much assessed as BM25 as effective and general approach).  Clearly MT errors can impact the quality of our candidates. However, if a question is badly translated the retrieved content will be inconsistent with the candidates retrieved for the question in the original language (and also inconsistent with candidates retrieved using questions translated in other languages). Our joint modeling through large generation-based Transformers can recover from these random errors.
For example, for 3 languages out of 5, we show that the CrossGenQA pipelines that use MT for the question outperform monolingual pipelines (\MonoGenQAshort and \MultiGenQAshort). This shows that translation errors are recovered by our approach.

\subsection{AWS-Translation for Machine Translation}
\label{sec:translate_aws}

For translating the questions automatically, we use AWS Translate. AWS Translate is a machine translation API that competes and outperforms in some cases other available translation APIs\footnote{cf. https://aws.amazon.com/blogs/machine-learning/amazon-translate-ranked-as-1-machine-translation-provider-by-intento/}. We compare the performance of a strong baseline on the FLORES dataset in table~\ref{tab:aws_translate_score}. We find that AWS translate outperforms the baseline for all the language pairs we work with. We leave for future work the study of the impact of different machine translation systems on our \CrossGenQA models. 

\section{Ethics Statement}

\subsection{Potential Harms of \GenQA}

All our \GenQA are fine-tuned from a large pretrained language model, \MTFIVE \cite{xue-etal-2021-mt5}. 
In general, large language models have been shown to have a potential to amplify societal biases \cite{bender2021parrots}, and might leak information about the datasets they were trained on \cite{carlini2021extracting}. 
In particular, the Colossal Cleaned Crawled Corpus (C4) and its multilingual counterpart (\textsc{mC4}) that were used to train \MTFIVE have been shown to disproportionately under-represent content about minority individuals \cite{dodge-etal-2021-documenting}.

In its use as a retrieval-based question answering system, \GenQA also can also cause harm due to (\textit{i}) the use of candidate sentences that are extracted from web documents, and (\textit{ii}) model hallucinations that are produced during decoding. In this work, (\textit{i}) is mitigated by only relying on content from Wikipedia, which, while not immune to vandalism \cite{alkharashi2018vandalism}, is of much higher quality of unvetted web data.
Regarding the risk of model hallucinations, this work does not attempt to directly mitigate any potential issue through modeling; rather, we always show annotators reference answer so that hallucination that result in factually incorrect answers can be properly caught during evaluation. 

\subsection{\GenTyDiQA Copyright}

Our \GenTyDiQA dataset is based on the TyDiQA dataset questions \citep{clark-etal-2020-tydi}. TyDiQA is released under the Apache 2.0 License which allows modification and redistribution of the derived dataset. Upon acceptance of this paper, we will release \GenTyDiQA and honor the terms of this license.

\GenTyDiQA answers were collected using Amazon Mechanical Turk. No geolocation filters or any personal information were used to hire turkers. Additionally, \GenTyDiQA questions treat scientific or cultural topics that can be answered objectively using Wikipedia. For these reasons, the collected answers cannot be used to identify their authors. 
Finally, to ensure the complete anonymity of the turkers, we will not release the turkers id along with the collected answers.

\subsection{Energy Consumption of Training}

All our experiments are based on the \MTFIVE base model. We run all our fine-tuning and evaluation runs using 8 Tesla P100 GPUs\footnote{\url{https://www.nvidia.com/en-us/data-center/tesla-p100/}}, which have a peak energy consumption of 300W each. Fine-tuning our \CrossGenQA models on MS-MARCO~\citep{DBLP:conf/nips/NguyenRSGTMD16} takes about 24 hours. 

\section{Reproducibility}
\subsection{Mechanical-Turk Settings}
\label{sec:mturk}

In this paper, we rely on Amazon Mechanical Turk for two distinct uses. 

On the one hand, we use it to build the \GenTyDiQA dataset.  For data collection, we request 1 turker per question to generate an answer. For the \GenTyDiQA data validation, we request 5 turkers to select only answers that are correct, aligned with the provided passage, self-contained and complete. 

On the other hand, we use Amazon Mechanical Turk to estimate the answer accuracy of our models. To do so, for each question, we provide the \GenTyDiQA reference and ask 3 turkers to vote on whether the generated answer is correct or not.

For those two uses, we use the following Amazon Mechanical Turk filters to hire turkers. 
\begin{itemize}
    \item We hire turkers that received at least a 95\% HIT\footnote{A HIT, as defined in Amazon Mechanical Turk, is a \textit{Human Intelligent Task}. In our case, a HIT consists in generating, validating, or accepting an answer to a single question.} approval rate.
    \item We request turkers that have performed at least 500 approved HITs.
    \item When possible, we use the ``\textit{master turker}'' filter\footnote{As stated on the Amazon Mechanical Turk website, "Amazon Mechanical Turk has built technology which analyzes Worker performance, identifies high performing Workers, and monitors their performance over time. Workers who have demonstrated excellence across a wide range of tasks are awarded the Masters Qualification. Masters must continue to pass our statistical monitoring to retain the Amazon Mechanical Turk Masters Qualification."} provided by Amazon Mechanical Turk. We find that this filter can only be used for English. For other languages, this filter leads to a too-small turker pool making it unusable in practice.

\end{itemize}

\begin{table}[t!]
\footnotesize\small 
\renewcommand*{\arraystretch}{1.15}
\centering
\begin{tabular}{lrr}
\toprule
  Parameter & Value  & Bounds\\
  \hline
  Effective Batch Size & 128 & [1, 8192] \\
  Optimizer & Adam  & -\\
  Learning Rate &  5e-4 & [1e-6,1e-3] \\
  Gradient Clipping value &  1.0 & - \\
  Epochs (best of) & 10  & [1, 30] \\
  Max Sequence Length Input & 524 & [1, 1024] \\
  Max Sequence Length Output  & 100 & [1, 1024] \\
\bottomrule
\end{tabular}
\caption{Optimization Hyperparameter to fin-tune \MTFIVE for the \GenQA task. For each hyper-parameter, we indicate the value used as well as the parameter lower and upper bounds when applicable.
}
\label{tab:hyperparameters}
\end{table}

\begin{table}[t!]
\centering\small
\resizebox{\linewidth}{!}{
\begin{tabu}{l@{\hspace{0.35cm}}cc}%
	\toprule
	Language & 
	\# Candidates & \% Positive Candidates\\
	\midrule
	\textsc{Ar} & 1,163,407 / 100,066  & 1.30 /  1.46\\ %
	\textsc{En} &   688,240 / 197,606    & 0.56 / 0.49\\ %
	\textsc{Bn} &  334,522 / 23892  & 0.76 / 0.74\\ %
	\textsc{Ja} &  827,628 / 214,524 & 0.47 / 0.47\\ %
	\textsc{Ru} & 1,910,388 / 245,326  & 0.34 / 0.48\\ %
	\bottomrule
\end{tabu}}
\caption{AS2-TyDiQA dataset extracted from the TyDiQA dataset. We report Train/Dev set following the TyDiQA split. We note that each question have at least one positive candidate}
\label{tab:as2tydiqa}
\end{table}

On Mechanical turk, the reward unit for workers is the HIT. In our case, a HIT is the annotation/validation of a single question. We make sure that each turker is paid at least an average of 15 USD/hour. To estimate the fair HIT reward, we first run each step with 100 samples ourselves in order to estimate the average time required per task. For data collection, we set the HIT reward to 0.50~USD based on an estimation of 0.5 HIT/min. For data validation, we set it to 0.15~USD based on an estimation of 1.6 HIT/min. For model evaluation, we set the HIT reward to  0.10~USD based on an estimation of 2.5~HIT/min. 

\subsection{Model Optimization}
All the \GenQA experiments we present in this paper are based on fine-tuning \MTFIVE base \citep{xue-etal-2021-mt5}. Models are implemented in PyTorch \citep{paszke2019pytorch}, and leverage \verb|transformers| \citep{wolf-etal-2020-transformers} and \verb|pytorch-lightning| \citep{falcon2020framework}. For fine-tuning, we concatenate the question and the candidate sentences, input it to the model and train it to generate the answer. Across all our runs, we use the hyperparameters reported in table~\ref{tab:hyperparameters}.

\section{Analysis}

\subsection{Gold vs. Retrieved Candidates}
We report in table~\ref{tab:multilingual_gen_QA} the performance of the \MonoGenQAshort and \MultiGenQAshort models when we feed them gold passages (using TyDiQA passage) and compare them with the performance of the same models fed with the retrieved candidates. We discuss those results in section \ref{subsec:analysis}.

\subsection{Human Evaluation vs. BLEU and ROUGE-L}
\label{sec:evaluation_bleu_vs_accuracy}

\begin{table}[t!]
\centering\footnotesize
\begin{tabu}{ l c c}%
	\toprule
	
    \textsc{Language}	& \textsc{w. BLEU}  & \textsc{w. ROUGE} \\
	\midrule
   \textsc{Ar} & 9.5 & 24.5 \\
   \textsc{Bn} & 21.2 & 5.3 \\
   \textsc{En} &11.7 & 23.5 \\
   \textsc{Ru} & 5.9 & 16.8 \\
	\bottomrule
\end{tabu}
\caption{Spearman Rank Correlation (\%) of human estimated Accuracy with BLEU and the ROUGE-L F~score. We run this analysis at the sentence level on the \MultiGenQA predictions. 
}%
\label{tab:speaman_accuracy_vs_others}
\end{table}

\begin{table}[t!]
\centering\footnotesize
\begin{tabu}{ l c c}%
	\toprule
	
    \textsc{Language}	& \textsc{w. BLEU}  & \textsc{w. ROUGE}
	\\
	\midrule
   \textsc{Ar} & 30.0 & 30.0\\
   \textsc{Bn} & -50.0 & -50.0\\
   \textsc{En} &40.0 & 40.0 \\
   \textsc{Ja} &-90.0 & -60.0 \\ 
   \textsc{Ru} & -87.2 & 100.0 \\

	\bottomrule
\end{tabu}
\caption{Spearman Rank Correlation (\%) of human estimated Accuracy with the BLEU score and the ROUGE-L F score at the model level across our 5 models (AS2, \MonoGenQAshort, \MultiGenQAshort, \CrossGenQAshort (x2)) 
}%
\label{tab:speaman_inter_models}
\end{table}

For comparison with previous and future work, we report the \textsc{BLEU} score (computed with SacreBLEU \citep{post-2018-call}) and the F-score of the \textsc{ROUGE-L} metric \citep{lin-2004-rouge} along with the human evaluation accuracy in table~\ref{tab:cross_lingual_candidates_w_BLEU_ROUGE}. 

As seen in previous work discussing the automatic evaluation of QA systems by \citet{Chaganty2018ThePO} and \citet{chen-etal-2019-evaluating}, we observe that for many cases, BLEU and ROUGE-L do not correlate with human evaluation. In table~\ref{tab:speaman_accuracy_vs_others}, we take the predictions of our \MultiGenQAshort model across all the languages and compute the Spearman rank correlation at the sentence level of the human estimated accuracy with BLEU and ROUGE-L. We find that this correlation is at most 25\%. This suggests that those two metrics are not able to discriminate between correct predictions and incorrect ones.

Additionally, we report the Spearman rank correlation between the Accuracy and BLEU or ROUGE across all our 5 models in table~\ref{tab:speaman_inter_models}. We find that neither BLEU nor ROUGE-L correlates strongly with human accuracy across all the languages. This means that those metrics are not able to rank the quality of a model in agreement with human judgment. Those results lead us to focus our analysis and to take our conclusions only on human evaluated accuracy. We leave for future work the development of an automatic evaluation method for multilingual \GenQA.

\begin{table*}[ht!]
\centering\footnotesize
\begin{tabu}{ l   @{\hspace{0.35cm}} l  l c  c c}%
	\toprule
	
  \bf  \textsc{Model}	& \bf\textsc{Question} & \bf\textsc{Candidates} & \bf BLEU   & \bf ROUGE & \bf Accuracy
	\\
	\midrule

    AS2 & \textsc{Ar} & \textsc{Ar}  &5.9 & 20.6 &   68.0 \\
	\MonoGenQAshort &\textsc{Ar}  &\textsc{Ar} &   17.2  &   38.8 & 68.4  \\ %
	\MultiGenQAshort  & \textsc{Ar} &\textsc{Ar} &   17.4 & 39.0  & 72.7 \\ %
	\CrossGenQAshort & \textsc{Ar} & \textsc{Ar-Bn-En-Ja-Ru Top 10}  & 15.3 & 36.5 & 72.0 \\  
	\CrossGenQAshort & \textsc{Ar} & \textsc{Ar-Bn-En-Ja-Ru Top 2 per lang.} & 14.7 & 36.3 & \bf 73.2  \\  

	\midrule

    AS2 & \textsc{Bn} & \textsc{Bn} &  3.8 & 16.6 & 58.0 \\ %
    \MonoGenQAshort & 	\textsc{Bn}  & \textsc{Bn}&     21.7 &   43.0 &  67.4 \\
    \MultiGenQAshort & 	\textsc{Bn}  &  \textsc{Bn} &     23.7  &   44.9 & \bf 76.5 \\ %
	\CrossGenQAshort & \textsc{Bn} &\textsc{Ar-Bn-En-Ja-Ru Top 10} &  35.2   & 56.5 & 25.3 \\  
	\CrossGenQAshort & \textsc{Bn} & \textsc{Ar-Bn-En-Ja-Ru Top 2 per lang.} & 33.5    & 54.8 & 18.5 \\

	\midrule
	
	AS2 & \textsc{En} &\textsc{En} &  5.6  & 20.0  & 39.0 \\ %
	\MonoGenQAshort & \textsc{En} & \textsc{En} & 23.0 &  46.4  &  \bf 43.6 \\
	\MultiGenQAshort & \textsc{En}  & \textsc{En} & 21.8 &  46.2  &  37.4 \\
	\CrossGenQAshort & \textsc{En} &\textsc{Ar-Bn-En-Ja-Ru Top 10}&    21.0 &  45.5	 & 31.0 \\
	\CrossGenQAshort & \textsc{En} & \textsc{Ar-Bn-En-Ja-Ru Top 2 per lang.} &   20.2   & 44.8 & 29.3 \\ 
	\midrule

	AS2 & \textsc{Ja}  & \textsc{Ja} & 6.7  & 22.4 &  70.4      \\ %
	
	\MonoGenQAshort & \textsc{Ja} & \textsc{Ja} &    19.4  & 45.0  & 64.3 \\  
	\MultiGenQAshort  & \textsc{Ja} & \textsc{Ja} &   19.1  &  45.5 & 65.5 \\
	\CrossGenQAshort & \textsc{Ja} & \textsc{Ar-Bn-En-Ja-Ru Top 10} &  17.6  & 42.2 &   70.3    \\ %
	\CrossGenQAshort & \textsc{Ja} &\textsc{Ar-Bn-En-Ja-Ru Top 2 per lang.}& 16.6    & 43.0 & \bf 71.6       \\ %

	\midrule
	AS2 & \textsc{Ru} & \textsc{Ru}  & 7.4  & 13.3 &  60.8\\
	\MonoGenQAshort &\textsc{Ru}  &  \textsc{Ru} &  6.4 &  23.4 & 61.3  \\ %
	\MultiGenQAshort  & \textsc{Ru} &\textsc{Ru} &   6.4  &   23.2   &  66.7 \\%
	\CrossGenQAshort & \textsc{Ru} & \textsc{Ar-Bn-En-Ja-Ru Top 10}   &4.2 & 21.0 & \bf 74.3  \\  
	\CrossGenQAshort & \textsc{Ru} & \textsc{Ar-Bn-En-Ja-Ru Top 2 per lang.} & 5.3 & 22.8   & \bf 74.7\\  

	\bottomrule
\end{tabu}
\caption{Performance of \GenQA models on \GenTyDiQA based on retrieved and reranked candidates. \textsc{Question} indicates the language of the question and the answer while \textsc{Candidates} indicates the language set of the retrieved candidate sentences. 
}%
\label{tab:cross_lingual_candidates_w_BLEU_ROUGE}
\end{table*}

\end{document}